\documentclass[]{template}

\usepackage[utf8]{inputenc}             
\usepackage[T1]{fontenc}                
\usepackage{url}                        
\usepackage{booktabs}                   
\usepackage{multirow}
\usepackage{colortbl}
\usepackage{multicol}
\usepackage{amsfonts}                   
\usepackage{nicefrac}                   
\usepackage{microtype}                  
\usepackage[dvipsnames]{xcolor}         

\usepackage{latexsym}

\usepackage{graphicx}
\usepackage{float}
\usepackage{subcaption}
\usepackage{wrapfig}
\usepackage{lipsum}

\usepackage{bm}

\usepackage{tabularx} 
\usepackage{ragged2e} 
\newcolumntype{L}{>{\RaggedRight\hangafter=1\hangindent=0em}X}

\usepackage{enumitem}

\usepackage{amsmath}
\usepackage{amssymb}
\usepackage{mathtools}
\usepackage{amsthm}
\usepackage{multicol}
\usepackage{multirow}
\usepackage{bbding}
\usepackage[table]{xcolor}
\usepackage{amsmath}
\usepackage{graphicx}
\usepackage{makecell}

\setboolean{logo}{true} 

\usepackage[linesnumbered,ruled,vlined]{algorithm2e}

\hypersetup{
    colorlinks=true,
    linkcolor=red,
    citecolor=Cerulean,
    filecolor=magenta,      
    urlcolor=magenta,
}

\usepackage[capitalize,noabbrev]{cleveref}
\crefname{section}{§}{§§}
\Crefname{section}{§}{§§}

\usepackage{calligra}
\DeclareMathAlphabet{\mathcalligra}{T1}{calligra}{m}{n}

\usepackage{pifont}

\theoremstyle{plain}

\theoremstyle{definition}

\theoremstyle{remark}

\renewcommand{\paragraph}[1]{\vspace{1mm}\noindent\textbf{#1}}

\DeclareCaptionLabelFormat{cont}{#1~#2\alph{ContinuedFloat}}
\usepackage[most]{tcolorbox}
\tcbset{
  promptbox/.style={
    top=10pt,
    colback=lightgray!20,
    colframe=Black,
    colbacktitle=NavyBlue,
    enhanced,
    center,
    attach boxed title to top center={yshift=-0.1in,xshift=0.0in},
    boxed title style={boxrule=0pt,colframe=white,},
  }
}
\newtcolorbox{promptbox}[2][]{promptbox, title=#2,#1}
\tcbset{
  takeawaybox/.style={
    top=10pt,
    colback=lightgray!20,
    colframe=Black,
    colbacktitle=BurntOrange,
    enhanced,
    center,
    attach boxed title to top center={yshift=-0.1in,xshift=0.0in},
    boxed title style={boxrule=0pt,colframe=white,},
  }
}
\newtcolorbox{takeawaybox}[2][]{takeawaybox, title=#2,#1}
\tcbset{
  observationbox/.style={
    top=10pt,
    colback=lightgray!20,
    colframe=Black,
    colbacktitle=YellowGreen,
    enhanced,
    center,
    attach boxed title to top center={yshift=-0.1in,xshift=0.0in},
    boxed title style={boxrule=0pt,colframe=white,},
  }
}
\newtcolorbox{observationbox}[2][]{observationbox, title=#2,#1}

\usepackage{xspace}

\newcommand\blfootnote[1]{%
  \begingroup
  \renewcommand\thefootnote{}\footnote{#1}%
  \addtocounter{footnote}{-1}%
  \endgroup
}

\usepackage{CJK}
\newcommand{\methodname}{ETCHR\xspace}
\title{ETCHR: Editing To Clarify and Harness Reasoning}

\author[1,2]{Beichen Zhang$^*$}
\author[2,3]{Yuhong Liu$^*$}
\author[1,2]{Jinsong Li}
\author[2]{Yuhang Zang$^\dagger$}
\author[4]{Jiaqi Wang$^\dagger$}
\author[1,2,5]{Dahua Lin$^\dagger$}

\affil[1]{The Chinese University of Hong Kong}
\affil[2]{Shanghai AI Laboratory}
\affil[3]{Shanghai Jiao Tong University}
\affil[4]{Shanghai Innovation Institute}
\affil[5]{CPII under InnoHK}

\begin{abstract}
Multimodal Large Language Models have advanced visual reasoning, yet a purely textual chain of thought remains a bottleneck for questions that require fine-grained focus or view transformations. The ``think with images'' paradigm narrows this gap, but existing approaches are either constrained by fixed predefined toolkits or produce noisy intermediate images from unified multimodal methods. We pursue a third option: using a dedicated image editing model and decouple it with an understanding model. However, off-the-shelf image editors fail as reasoning assistants with two complementary gaps: a \emph{language-side} gap, where editors trained as passive instruction-followers cannot map an abstract question to an appropriate visual transformation, and a \emph{generation-side} gap, where edit correctness degrades as reasoning depth grows. Guided by this analysis, we introduce \textbf{ETCHR} (\textbf{E}diting \textbf{T}o \textbf{C}larify and \textbf{H}arness \textbf{R}easoning), a question-conditioned, reasoning-aware image editor decoupled from the downstream understanding model and trained with a two-stage recipe targeted at the two gaps: Reasoning Imitation via supervised fine-tuning on edit trajectories, followed by Reasoning Enhancement with VLM-derived rewards for edit correctness and downstream reasoning accuracy. Since the editor is decoupled, ETCHR plugs into different open- and closed-source MLLMs in a training-free manner. Across five task families (fine-grained perception, chart understanding, logic reasoning, jigsaw restoration, and 3D understanding), ETCHR raises average Pass@1 from 55.95 to 60.77 (+4.82) with Qwen3-VL-8B, from 65.08 to 70.55 (+5.47) with Gemini-3.1-Flash-Lite, and from 76.55 to 81.16 (+4.61) with the 1T-parameter MoE model Kimi K2.5.
\end{abstract}

\begin{document}

\blfootnote{$\dagger$ Corresponding authors: zangyuhang@pjlab.org.cn}
\blfootnote{$*$ Eqaul Contribution. Code is at \url{https://github.com/InternLM/ETCHR}}

\maketitle

\section{Introduction}

\begin{figure}[t]
  \centering
    \includegraphics[width=\linewidth]{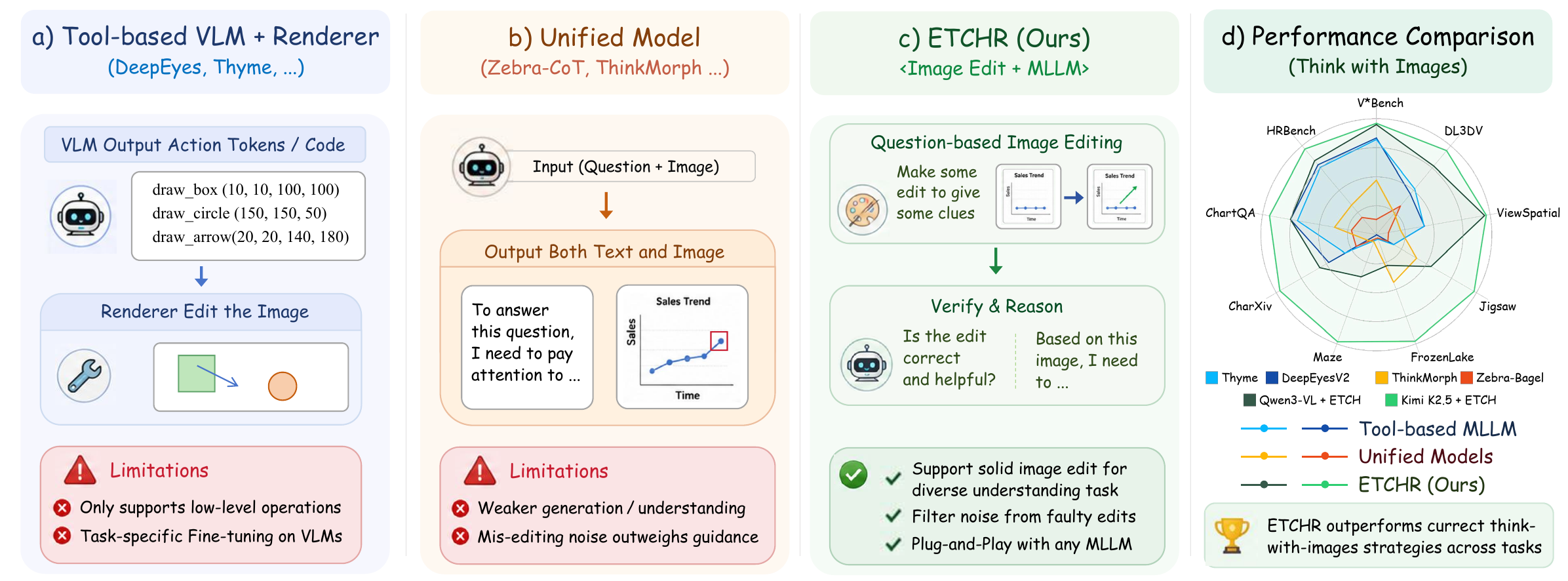}
\caption{\textbf{\methodname vs.\ prior ``think with images'' paradigms.}
\textbf{(a)} \emph{Tool-based} methods emit action tokens to a renderer, limiting edits to low-level operations and requiring VLM fine-tuning.
\textbf{(b)} \emph{Unified models} share one backbone for text and images, weakening both and producing noisy intermediates.
\textbf{(c)} \methodname decouples a question-conditioned editor from the understanding MLLM and adds a verify-and-reason step, enabling plug-and-play use across tasks.
\textbf{(d)} Across nine benchmarks, \methodname (with Qwen3-VL-8B and Kimi K2.5 1T) surpasses tool-based and unified-model baselines.}
\label{fig:teaser}
\end{figure}

Recent Multimodal Large Language Models (MLLMs)~\cite{hurst2024gpt,comanici2025gemini,bai2025qwen3} have substantially improved visual reasoning~\cite{yue2024mmmu,lu2023mathvista}, but a purely textual chain-of-thought~\cite{wei2022chain} remains a bottleneck when a question depends on \textit{where} to look or \textit{how} a scene would change under an action. In such cases, the model must verbalize a spatial state it cannot draw, and small descriptive errors compound across steps. A growing line of work therefore enables MLLMs to \textbf{``think with images''}~\cite{openai2025thinkingwithimages,zheng2025deepeyes,hu2024visual}, generating an intermediate image during inference and feeding it back into the reasoning trajectory, with reported gains on fine-grained visual search~\cite{wu2024v,zheng2025deepeyes,hu2024visual}, chart reasoning~\cite{zhang2025thyme,hong2025deepeyesv2}, and spatial navigation~\cite{gu2025thinkmorph,li2025zebra}.

The difficulty of the ``think with images'' paradigm is not simply generating an image, but generating the \emph{right} intermediate image for the question. The system must infer what visual change would advance the answer, render that change faithfully, and produce an intermediate image that the downstream understanding model can exploit. This couples two capabilities that are hard to satisfy simultaneously: \textbf{(i)} the \emph{breadth} of visual transformations the system can conceive and execute, ranging from fine-grained highlight of local elements to holistic transitions in spatial perspective; and \textbf{(ii)} the \emph{fidelity} with which those transformations are rendered, ensuring the intermediate image can indeed serve as reliable evidence for downstream reasoning. Only when both capabilities are present can a think-with-image framework deliver high-quality assistance across diverse understanding tasks.

Existing approaches address only one side. \textit{Tool-based} methods (Fig.~\ref{fig:teaser} \textbf{(a)})~\cite{hu2024visual,wu2024v,zheng2025deepeyes,hong2025deepeyesv2,zhang2025thyme} fine-tune the understanding model to emit bounding boxes, crop/zoom commands, or executable snippets that a deterministic renderer applies; the actions are controllable but confined to low-level, localized manipulations, and task-specific fine-tuning can erode the model's general competence~\cite{kirkpatrick2017overcoming}. \textit{Unified multimodal} methods (Fig.~\ref{fig:teaser} \textbf{(b)})~\cite{team2024chameleon,chern2024anole,deng2025emerging,gu2025thinkmorph,li2025zebra} instead use a single backbone to interleave text and image tokens, gaining flexibility but inheriting a weaker generative head: recent work~\cite{wen2026unig2u} shows that their intermediate images often inject noise rather than guidance, while the unified backbone itself lags specialist understanding and generation models~\cite{wu2025janus,xie2024show}. Both families share a further blind spot: \textbf{neither verifies whether the intermediate edit is actually correct} before reasoning forward from it~\cite{madaan2023self,shinn2023reflexion,huang2023large}, allowing a noisy edit to propagate directly into the final answer.

These limitations motivate us to pursue a \textbf{third option}: using a dedicated image-to-image editor as the intermediate-image generator. A specialist editor expresses a far broader transformation space than the predefined actions of tool-based methods~\cite{hu2024visual,wu2024v,zheng2025deepeyes,hong2025deepeyesv2,zhang2025thyme}, preserves the fidelity of a model dedicated to editing rather than the weaker generative head of a unified backbone~\cite{team2024chameleon,chern2024anole,deng2025emerging,gu2025thinkmorph,li2025zebra}, and decouples edit quality from the understanding model so the latter need not be retrained on task-specific editing formats. This option is now plausible architecturally: modern image-to-image editors (e.g., FLUX-class~\cite{blackforestlabs2025flux2} and Qwen-Image-Edit~\cite{wu2025qwen}) replace the shallow CLIP-style text encoder~\cite{radford2021learning} of earlier editors with an MLLM-style encoder, giving the editor enough language-side capacity to parse a complex question and support cross-task visual transformations within a single model.

Realizing this option, however, is non-trivial. Image editors are normally trained as \textit{passive instruction-following tools} that expect an explicit edit prompt such as ``add a bounding box around the red car''~\cite{brooks2023instructpix2pix,zhang2023magicbrush}, whereas a model that thinks with images is handed only a \emph{question} and must itself decide what edit would help answer it. Moreover, the generative components of current editing models often lack the capacity for sophisticated reasoning, making it difficult to assist those understanding tasks requiring complex logic or 3D spatial manipulations. The \textbf{central challenge} is therefore to turn the editor into an autonomous question-conditioned system with strong reasoning capabilities, whose intermediate images are genuinely reasoning-useful rather than merely visually plausible. Correctness is also critical: an erroneous edit can silently mislead the downstream understanding model without verification~\cite{madaan2023self,shinn2023reflexion,huang2023large}, so the system must also be able to detect and reject its own bad edits before they enter the reasoning trajectory.

We instantiate this design as \textbf{\methodname} (\textbf{E}diting \textbf{T}o \textbf{C}larify and \textbf{H}arness \textbf{R}easoning, (Fig.~\ref{fig:teaser} \textbf{(c)})): a \textbf{question-conditioned, reasoning-aware image editor} coupled with edit-verification at inference. This design uses question-conditioned editing to strengthen \textit{language-side reasoning} and a dedicated specialist editor to preserve \textit{generative fidelity}. Unlike \textit{tool-based} methods~\cite{hu2024visual,wu2024v,zheng2025deepeyes,hong2025deepeyesv2,zhang2025thyme}, \methodname is not confined to a predefined action space and infers what to change directly from the question. Unlike \textit{unified multimodal} methods~\cite{team2024chameleon,chern2024anole,deng2025emerging,gu2025thinkmorph,li2025zebra}, \methodname keeps understanding and generation in separate specialist models, so editing fidelity is not compromised by joint optimization. Going beyond both families, \methodname incorporates a \textit{reflective inference step} at inference, letting the understanding model reject noisy edits rather than propagate them to the final answer.

\methodname is trained in two stages and deployed with a reflective inference procedure. (1) \textbf{Reasoning Imitation} applies supervised fine-tuning to convert the editor from a passive renderer into a question-conditioned editor that infers the useful transformation from the question alone. (2) \textbf{Reasoning Enhancement} then uses reinforcement learning with \textit{reasoning-aware rewards}, rather than only visual plausibility, to push outputs toward edits that are correct in isolation and useful for downstream reasoning. At inference, \methodname follows an \textbf{Edit-Verify-Reason} procedure: the editor proposes an intermediate image, the understanding model verifies whether the proposed edit is reliable, and the answer is produced from the edited image only when verification succeeds; otherwise, the system falls back to the original image.

On a diverse benchmark suite covering fine-grained perception (V*Bench~\cite{wu2024v}, HRBench~\cite{hrbench}), chart understanding (ChartQA~\cite{masry-etal-2022-chartqa}, CharXiv~\cite{wang2024charxiv}), logic and path reasoning (Maze, Frozen Lake), jigsaw reasoning (built on COCO~\cite{Lin2014MicrosoftCC}), and 3D understanding (ViewSpatial~\cite{Li2025ViewSpatialBenchEM}), \methodname raises average Pass@1 from 55.95 to 60.77 (+4.82) with Qwen3-VL-8B~\cite{bai2025qwen3}, from 65.08 to 70.55 (+5.47) with Gemini-3.1-Flash-Lite and from 76.55 to 81.16 (+4.61) with Kimi K2.5, outperforming tool-based~\cite{hong2025deepeyesv2,zhang2025thyme} and unified-model~\cite{li2025zebra,gu2025thinkmorph} baselines (Fig.~\ref{fig:teaser} \textbf{(d)}).
Ablations confirm that both training stages, the two reward signals (editing correctness and editing guidance), and the Edit-Verify-Reason reflection at inference, each contribute and are complementary.

Our contributions:
\textbf{1)} We propose a mechanism for ``thinking with images'' that uses a dedicated image-to-image editor as the source of intermediate visual evidence, with an inference-time verification step that lets the understanding model reject unreliable edits.
\textbf{2)} We instantiate this mechanism as \textbf{\methodname}, a reasoning-aware editor built on FLUX.2-klein-base-9B~\cite{blackforestlabs2025flux2} and trained with a two-stage recipe: \textit{Reasoning Imitation}, supervised fine-tuning on question-conditioned edit trajectories, followed by \textit{Reasoning Enhancement} with two VLM-derived reward signals.
\textbf{3)} Because \methodname conditions on the question and is decoupled from the understanding model, it pairs with different open- and closed-source MLLMs without fine-tuning them, evaluated across reasoning tasks spanning fine-grained perception, spatial and path reasoning, puzzle restoration, and 3D understanding.

\section{Analysis}
\label{sec:analysis}

\begin{figure}[t]
  \centering
    \includegraphics[width=\linewidth]{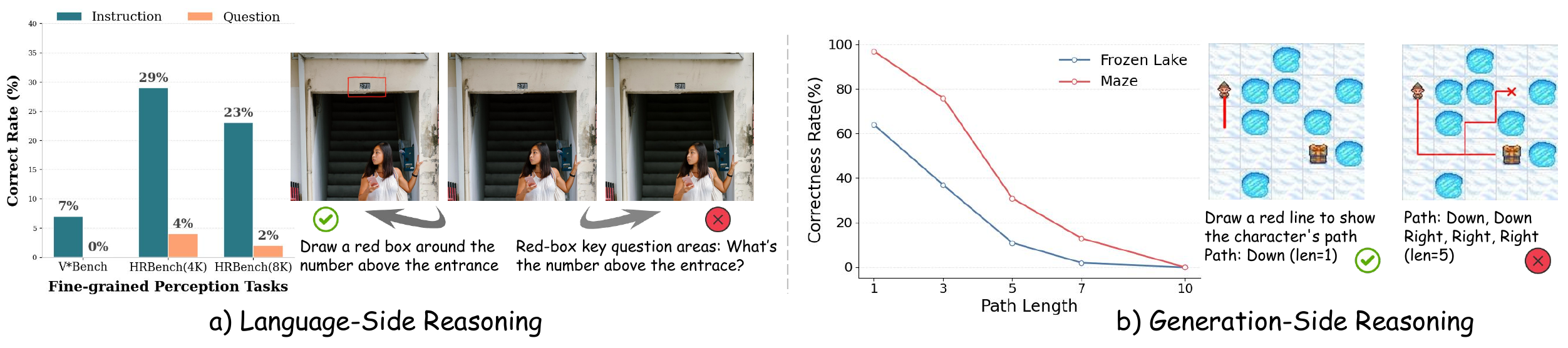}
    \caption{\textbf{Diagnosis of two complementary gaps in image editors.} \textbf{(a)} \textbf{Language-side reasoning}: editors trained as instruction-followers perform worse on abstract questions than on concrete instructions, revealing a gap in mapping reasoning questions to visual transformations. \textbf{(b)} \textbf{Generation-side reasoning}: even with concrete instructions, edit correctness decreases as path length $L$ grows on Maze Solving, indicating difficulty executing multi-step reasoning during DiT decoding.}
    \label{fig:analysis}
\end{figure}

To motivate \methodname, we decompose reasoning-aware editing into two sub-capabilities that current editors lack. \emph{Language-side reasoning} infers, from an abstract question alone, what visual transformation would help answer it. \emph{Generation-side reasoning} faithfully renders that transformation when it requires non-trivial spatial or algorithmic inference (e.g., tracing a maze path). We diagnose both gaps with off-the-shelf editors and understanding models.

\noindent \textbf{Language-Side Reasoning: From Question to Edit.}
Modern image-to-image editors (e.g., FLUX.2-klein-base-9B~\cite{blackforestlabs2025flux2}, InstructPix2Pix~\cite{brooks2023instructpix2pix}) are optimized as instruction-following tools: they expect an explicit edit prompt such as ``draw a red box around the trash can'' rather than an abstract question such as ``Is the trash can on the left or right side of the black chair?'' We therefore probe a question-to-edit gap: whether the editor can recover the useful transformation from the raw question alone.

We use Gemini-3.1-Flash-Lite~\cite{comanici2025gemini} as a prompt enhancer that converts each question $q$ into a concrete editing instruction $p_{inst}$, and Qwen3-VL-8B~\cite{bai2025qwen3} judges correctness on 100 samples from V*Bench~\cite{wu2024v} and HRBench~\cite{hrbench}. As shown in Fig.~\ref{fig:analysis} (a), the concrete-instruction condition significantly outperforms the abstract-question condition. Shaped by explicit instruction-following, the base editor lacks a reliable question-to-edit mapping; reasoning-aware editing therefore needs more than generic image-editing fidelity.

\noindent \textbf{Generation-Side Reasoning: Scaling Robustness to Reasoning Depth.}
Despite efforts to address the weak language-side reasoning with prompt enhancers, we argue that their reasoning ability is even more lacking during the DiT decoding phase. Even with a concrete instruction, an editor can fail when the transformation itself requires multi-step spatial reasoning during DiT decoding. We measure edit correctness as a function of task complexity on Maze Solving and Frozen Lake Solving.

For these two tasks, we construct a held-out set of 100 samples with shortest-path lengths $L\in\{1,3,5,7,10\}$ and present the shortest path in the text prompt. A VLM-as-Judge marks an edit correct only if the highlighted path matches the input. Fig.~\ref{fig:analysis}~(b) plots edit correctness against $L$: accuracy is near-perfect at $L{=}1$ and decreases sharply as $L$ grows, approaching zero on the longest paths. This indicates that even when instructions are provided with maximum precision and granularity, current editing models still struggle to execute operations requiring multi-hop reasoning during the DiT decoding phase.

\noindent \textbf{Summary.} Together, the two gaps motivate \methodname's two-stage design: Stage~I question-conditioned imitation to close the question-to-edit gap and equip with strong reasoning capabilities, and Stage~II reasoning-oriented enhancement to further enhance multi-step generation.

\section{\methodname: Editing To Clarify and Harness Reasoning}
\methodname trains a question-conditioned image editor in two stages. After preliminaries (Sec.~\ref{sec:prelim}), \textbf{Stage~I (Reasoning Imitation, Sec.~\ref{sec:sft})} converts the base editor into a question-conditioned editor via supervised fine-tuning, and \textbf{Stage~II (Reasoning Enhancement, Sec.~\ref{sec:rl})} aligns it with downstream reasoning utility via RL under VLM-derived rewards. At inference, \textbf{Edit-Verify-Reason} (Sec.~\ref{sec:inference}) verifies each edit and reverts to the original image on failure.

\subsection{Preliminaries}
\label{sec:prelim}
We cast reasoning-aware editing as a question-conditioned image-to-image task. Each training instance is a tuple $\{i, i_{gt}, q, a\}$, where $i$ is the input image, $q$ is a natural-language question, $a$ is a ground-truth answer, and $i_{gt}$ is a ground-truth edited image that surfaces the visual evidence needed to derive $a$ from $i$ (e.g., a traced path through a maze). \methodname learns a question-conditioned editor $\mathcal{E}$ that maps $(i, q)$ to an edited image $i_{edit} = \mathcal{E}(i, q)$ serving the same evidential role as $i_{gt}$. Quality of $i_{edit}$ is judged by a frozen understanding model $\mathcal{M}$: a useful edit satisfies $\mathcal{M}(i, i_{edit}, q) = a$ while the unaided baseline satisfies $\mathcal{M}(i, q) \neq a$. This notation is reused throughout Sec.~\ref{sec:sft} and Sec.~\ref{sec:rl}.

\subsection{Reasoning Imitation Supervised Fine-tuning (SFT)}
\label{sec:sft}
\textbf{Data Preparation.} We build a large-scale SFT corpus of question-conditioned edit trajectories $\{i, i_{gt}, q, a\}$, partitioned into \textit{five reasoning families} spanning the visual transformations required by downstream reasoning tasks: \emph{fine-grained perception} probes localization of small or easily missed referents; \emph{chart understanding} probes grounding in structured plots; \emph{logic reasoning} probes multi-step algorithmic inference; \emph{jigsaw reasoning} probes global geometric reorganization; and \emph{3D understanding} probes viewpoint and camera-pose transformation. Our task coverage, from local annotation to whole-image rearrangement, prevents collapse onto a single edit template and forces the editor to acquire a \textit{meta-capability} of inferring what visual transformation each question demands.

For \emph{fine-grained perception}, we use the $V^*$ training dataset~\cite{wu2024v}, covering both $V^*$-GQA and $V^*$-COCO subsets, and synthesize $i_{gt}$ by rendering the annotated bounding boxes onto $i$. For \emph{chart understanding}, we draw from RefChartQA~\cite{vogel2025refchartqa} and apply the same bounding-box overlay procedure to obtain $i_{gt}$. For \emph{logic reasoning}, we build an in-house maze corpus where $i$ shows the maze topology and $i_{gt}$ overlays the correct traversal path. For \emph{jigsaw reasoning}, we sample from Spatial-SSRL~\cite{liu2025spatial}, taking $i$ as a spatially shuffled image and $i_{gt}$ as its restoration. For \emph{3D understanding}, we use DL3DV-10K~\cite{ling2024dl3dv}, which contains videos of real-world 3D scenes together with per-frame camera poses. We sample $i$ and $i_{gt}$ from the same video and synthesize $q$ and $a$ from the camera extrinsics.

\textbf{Task-level Prompt Enhancement.} Using $q$ alone as the editing prompt causes severe cross-task interference: the editor's latent space, shaped by explicit instruction-following, lacks the priors to disambiguate whether a question demands localization, path-tracing, rearrangement, or viewpoint transformation. We therefore prepend a task-level meta-prompt $p_{task}$ to $q$, evoking the editing modality appropriate to each task family.
At training, $p_{task}$ acts as a soft task-router that partitions the editor's latent space into task-specific manifolds and suppresses gradient conflicts across families. At inference, it adds no architectural cost: because $p_{task}$ is task-level, the editor needs no access to the understanding model's internal representations or per-instance instructions, letting \methodname be deployed in a training-free manner atop any open- or closed-source MLLMs.

\textbf{Training Strategy.} We use Low-Rank Adaptation (LoRA) \cite{hu2022lora} to fine-tune the Diffusion Transformer (DiT) \cite{peebles2023scalable} of FLUX.2-klein-base-9B \cite{blackforestlabs2025flux2}, keeping the VAE and text encoder frozen as inputs are standard frames and prompts are simply $p_{task}$ concatenated with $q$. We apply a large LoRA rank ($r = 768$) to all linear layers in the DiT blocks to provide sufficient capacity for multi-task learning.

\begin{figure}[t]
  \centering
    \includegraphics[width=\linewidth]{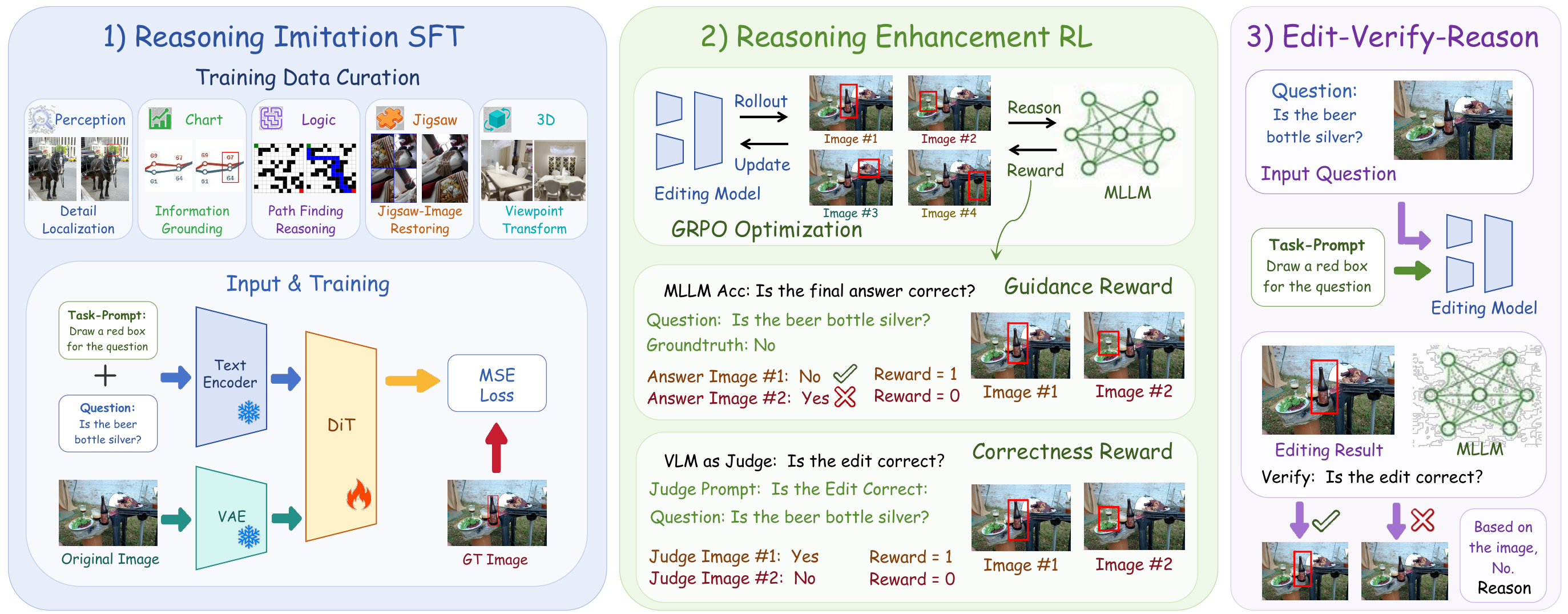}
    \caption{\textbf{Overview of \methodname.} \textbf{(1) Reasoning Imitation SFT} fine-tunes the base editor on curated question-edit trajectories, teaching it to infer the useful visual transformation from the question alone. \textbf{(2) Reasoning Enhancement RL} applies GRPO with two VLM-derived rewards, edit correctness and downstream answer accuracy, to align the editor with reasoning utility rather than visual plausibility. \textbf{(3) Edit-Verify-Reason} inference lets the understanding model verify each proposed edit and fall back to the original image when verification fails.}
    \label{fig:framework}
\end{figure}

\subsection{Reasoning Enhancement Reinforcement Learning (RL)}
\label{sec:rl}
\textbf{Data Preparation.} We curate the RL set by sampling 2{,}000 instances from each of the five families (fine-grained perception, chart understanding, logic reasoning, jigsaw, and 3D understanding) in the SFT corpus, yielding 10{,}000 pairs. An instance $(i, i_{gt}, q, a)$ is retained only if it satisfies:
\begin{equation}
\label{eq:rl-filter}
\mathcal{C}(i, i_{gt}, q, a) = \mathbf{1} [\mathcal{M}(i, q) \neq a \land \mathcal{M}(i, i_{gt}, q) =a],
\end{equation}
where the notation follows Sec.~\ref{sec:prelim}.
We keep only samples where (i) the understanding model fails on the raw image and (ii) succeeds when conditioned on the ground-truth edit. The first clause discards instances already solvable without visual assistance, avoiding wasted gradients on redundant edits; the second guarantees a verifiable upper-bound signal per sample, yielding a denser reward landscape and lower-variance policy gradients.

\textbf{Reward Design.}
Edit quality admits no direct scalar metric, so we infer it through two complementary rewards that probe distinct facets of edit adequacy and compensate for each other's blind spots.

\emph{Editing Guidance Reward ($r_{guide}$)}. The first reward measures the downstream reasoning utility of $i_{edit}$: it is one iff $\mathcal{M}$ answers correctly given $(i, i_{edit})$,
\begin{equation}
\label{eq:r-guide}
    r_{guide}(i, i_{edit}, q, a) = \mathbf{1}[\mathcal{M}(i, i_{edit}, q) = a].
\end{equation}

$r_{guide}$ is the most faithful signal, optimizing the end-to-end objective of producing edits that help $\mathcal{M}$ answer correctly. Its fidelity, however, is bounded by $\mathcal{M}$'s capability ceiling: on easy questions, $\mathcal{M}$ may succeed with a partially erroneous edit, and on hard ones it may fail even with a perfect edit. The filter $\mathcal{C}$ in Eq.~(\ref{eq:rl-filter}) reduces but cannot remove this coupling, motivating a second, decoupled signal.

\emph{Editing Correctness Reward ($r_{correct}$)}. To break this coupling, a second reward evaluates the edit \textit{in isolation}, without solving the underlying task. Following the VLM-as-Judge paradigm~\cite{zheng2023judging}, a judge VLM $\mathcal{J}$ assesses only whether $i_{edit}$ contains the visual information needed to answer $q$:
\begin{equation}
\label{eq:r-correct}
    r_{correct}(i, i_{edit}, q) = \mathbf{1}[\mathcal{J}(i, i_{edit}, q) = 1].
\end{equation}
$r_{correct}$ lifts the capability ceiling: $\mathcal{J}$ can recognize that a bounding box highlights the correct referent or an annotation traces the valid path, even without the reasoning depth to derive the answer, giving a broader notion of edit adequacy. The cost is judge noise: $\mathcal{J}$ may accept a plausible-but-uninformative edit, or reject a correct one due to superficial visual differences.

The two rewards are complementary: $r_{guide}$ is faithful but ceiling-bound, while $r_{correct}$ is ceiling-lifting but judge-noisy. We combine them as a convex sum:
\begin{equation}
\label{eq:reward-unified}
    \mathcal{R}(i, i_{edit}, q, a) = \alpha \, r_{guide} + \beta \, r_{correct},
\end{equation}
with $\alpha = \beta = 0.5$ by default, so that each term partially compensates for the other's blind spot.
To reduce variance, we soften both indicators into empirical probabilities: $\mathcal{M}$ and $\mathcal{J}$ are each queried with $K$ stochastic decodings, and $r_{guide}$, $r_{correct}$ are the fractions of correct answers and positive verdicts, respectively.

\textbf{Optimization.} We use Pref-GRPO~\cite{wang2025pref}, a pairwise-preference extension of GRPO~\cite{shao2024deepseekmath}. Given a rollout group of $G$ edited images $\{i^j_{edit}\}_{j=1}^G$ from the same $(i, q)$, we compute a pairwise win rate for each image under the combined reward $\mathcal{R}$. The win rate of image $m$ is:
\begin{equation}  
w_m = \frac{1}{G-1}\sum_{n \neq m}\mathbf{1}[\mathcal{R}(i, i_{edit}^m,q,a) > \mathcal{R}(i, i_{edit}^n,q,a)].
\end{equation}
Win rates are normalized within the group to give the advantage:
\begin{equation}
    \hat{A}_m = \frac{w_m - \mathrm{mean}(\{w_j\}_{j=1}^G)}{\mathrm{std}(\{w_j\}_{j=1}^G)}.
\end{equation}
By replacing absolute reward with pairwise preference, Pref-GRPO amplifies reward variance and absorbs reward noise, mitigating the illusory advantage problem and stabilizing training.

\subsection{Edit-Verify-Reason Inference}
\label{sec:inference}
Even a well-trained editor can produce misleading intermediates. To prevent such errors from contaminating reasoning, \methodname adds a reflective inference procedure that verifies edit reliability before use.
The pipeline has three steps: \textbf{(1) Edit.} The editor takes the concatenated prompt $[p_{task}, q]$ with image $i$ and produces a candidate edit $i_{edit}$. \textbf{(2) Verify.} $\mathcal{M}$ performs a binary check on whether $i_{edit}$ contains the visual information needed to answer $q$, without producing the answer itself.
\textbf{(3) Reason.} If verification passes, $\mathcal{M}$ conditions on both $i$ and $i_{edit}$ to produce the answer: $a = \mathcal{M}(i, i_{edit}, q).$
Otherwise, it falls back to the original image alone:
$a = \mathcal{M}(i, q).$
Verification is vital as edit errors are asymmetric in cost: a correct edit gives decisive visual guidance, while an incorrect one introduces structured confounders that MLLMs struggle to override.

\section{Experiments}

\textbf{Models.} We build our pipeline on top of multiple understanding models: the open-source Qwen3-VL-8B-Instruct~\cite{bai2025qwen3} (dense model with 8B parameters), Kimi K2.5~\cite{team2026kimi} (Mixture-of-Experts model with 1T parameters), and the closed-source Gemini-3.1-Flash-Lite-Preview \cite{comanici2025gemini}.

\noindent
\textbf{Benchmarks.} We evaluate our model on five categories of understanding tasks. \textbf{(1)} \textit{Fine-grained perception:} V*Bench~\cite{wu2024v}, which probes recognition of small details in high-resolution natural images, and HRBench (4K/8K)~\cite{hrbench}, which extends this setting to 4K and 8K inputs. \textbf{(2)} \textit{Chart understanding:} ChartQA~\cite{masry-etal-2022-chartqa} for question answering over bar, line, and pie charts, and CharXiv~\cite{wang2024charxiv} (Descriptive/Reasoning), which targets charts from arXiv papers. \textbf{(3)} \textit{Logical reasoning:} 200 Maze and 200 Frozen Lake tasks that we construct in-house, requiring the model to navigate from a start cell to a goal while respecting walls or hazards. \textbf{(4)} \textit{Jigsaw:} 200 puzzles built from MS COCO~\cite{Lin2014MicrosoftCC} images by applying tile-shuffling strategies, which probe whether the model can recover image structure from permuted patches. \textbf{(5)} \textit{3D perception and reasoning:} the Person-Perspective Relative-Direction subtask of ViewSpatial-Bench~\cite{Li2025ViewSpatialBenchEM}, together with DL3DV-2k~\cite{ling2024dl3dv}, a benchmark we constructed using scenes and images from DL3DV, comprising 2000 questions targeting the reasoning capability over 3D perspective-transformation.

\noindent
\textbf{Implementation Details.} We report accuracy (Pass@1, temperature 0) across all experiments. For baselines, we set the maximum number of tool invocations to 10 for structured-action models, and the maximum number of image generations to 10 for unified models.

\noindent \textbf{Training Details.}
For \textit{Reasoning Imitation Supervised Fine-tuning}, we use Low-Rank Adaptation (LoRA) to fine-tune the Diffusion Transformer of FLUX.2-klein-base-9B for 1 epoch with lora rank $r=768$, learning rate $lr = 1e^{-4}$, Classifier-free guidance $scale=1$, inference steps $step=30$. The supervised fine-tuning framework is built on DiffSynth-Studio.
For \textit{Reasoning Enhancement Reinforcement Learning}, we use Pref-GRPO to further fine-tune the DiT. The RL framework is built on Pref-GRPO~\cite{wang2025pref}. We also adapt LoRA fine-tuning strategy, with lora rank $r=128$, lora alpha $\alpha=128$, learning rate $lr=1e^{-4}$, Group Number for each rollout $G=8$. We use Qwen3-VL-8B-Instruct~\cite{bai2025qwen3} as both the understanding model $\mathcal{M}$ and judge model $\mathcal{J}$ to provide editing guidance reward and editing correctness reward.

\begin{table}[t]
    \caption{\textbf{Main results on five task families.} \methodname 
    is compared against tool-based and programmatic baselines and unified multimodal models. $\dagger$ marks tasks outside a baseline's supported image-editing scope; all numbers are Pass@1 at temperature 0.}
  \centering
  \setlength{\tabcolsep}{2.8pt}
  \scalebox{0.85}{
  \normalsize
  \begin{tabular}{@{}lcccccccccc@{}}
    \toprule
    \multirow{2}{*}{Models} & \multicolumn{2}{c}{Fine-grained Perception} & \multicolumn{2}{c}{Chart Understanding} & \multicolumn{2}{c}{Logic} & Jigsaw & \multicolumn{2}{c}{3D Understanding} &\multirow{2}{*}{\textbf{Avg.}}  \\
\cmidrule(lr){2-3}
\cmidrule(lr){4-5}
\cmidrule(lr){6-7}
\cmidrule(lr){8-8} \cmidrule(l){9-10} 
  & \makecell{\small V*Bench} & \makecell{\small HRBench\\ \scriptsize(4K / 8K)} & \small ChartQA & \makecell{\small CharXiv\\ \scriptsize(DQ / RQ)} &{\small Maze}& \makecell{\small Frozen\\ \small Lake}  & \makecell{\small COCO\\Jigsaw} & \makecell{\small ViewSpatial\\ \scriptsize Person-RelDir} & \small DL3DV-2k\\
    \midrule
    \rowcolor{blue!8}
    \multicolumn{11}{c}
    {\textit{Tool-based \& Programmatic Models}} \\
    \cmidrule(){1-11}DeepEyesV2 \cite{hong2025deepeyesv2} & 81.8 & 77.9 / 73.8 & 88.4 & 78.6 / 48.9 & $0.5^{\dagger}$ & $0.0^{\dagger}$ & $6.5^{\dagger}$ & $39.3^{\dagger}$ & $69.0^{\dagger}$ & \textbf{51.34}\\
    Thyme \cite{zhang2025thyme} & 81.2 & 77.1 / 71.9 & 86.5 & 66.1 / 45.3 & $0.0^{\dagger}$ & $2.0^{\dagger}$ & $6.5^{\dagger}$ & $39.2^{\dagger}$ & $70.8^{\dagger}$ & \textbf{49.69}\\
    \midrule
    \rowcolor{blue!8}
    \multicolumn{11}{c}{\textit{Unified-Models}} \\
    \cmidrule(){1-11}
    Bagel-Zebra-CoT \cite{li2025zebra} & 50.8 & 51.9 / 45.8 & 71.8 & 62.8 / 37.4 & 0.0 & 1.5 & 5.5 & 28.6 & 64.9 & \textbf{38.27}\\
    ThinkMorph-7B \cite{gu2025thinkmorph} & 65.8 & 59.6 / 50.9 & 76.4 & 61.2 / 28.1 & 6.5 & 31.0 & 10.5 & 32.2 & 62.4 & \textbf{44.05}\\
    \midrule 
    \rowcolor{blue!8}
    \multicolumn{11}{c}{\textit{Baselines \& Our Image Editing Model}} \\
    \cmidrule(){1-11}
    Qwen3-VL-8B \cite{bai2025qwen3} &  84.8& 78.1 / 74.5 & 86.7 & 85.0 / 47.9 & 27.5 & 9.5 & 9.5 & 41.1 & 70.8 & \textbf{55.95} \\
    +\methodname & 86.9 & 80.6 / 75.3 & 88.4 & 86.8 / 49.7 & 38.5 & 13.5 & 13.0 & 57.2 & 78.6 & \textbf{60.77} \\
    \rowcolor{gray!10}
    \textit{Improvement} & \textcolor[RGB]{0, 150, 50}{+2.1}  & 
    \textcolor[RGB]{0, 150, 50}{+2.5 / +0.8} &
    \textcolor[RGB]{0, 150, 50} {+1.7}&
    \textcolor[RGB]{0, 150, 50}{+1.8 / +1.8} &
    \textcolor[RGB]{0, 150, 50}{+11.0} &
    \textcolor[RGB]{0, 150, 50}{+4.0}&
    \textcolor[RGB]{0, 150, 50}{+3.5} & \textcolor[RGB]{0, 150, 50}{+16.1} & \textcolor[RGB]{0, 150, 50}{+8.2} &
    \textbf{\textcolor[RGB]{0, 120, 50}{\textbf{+4.82}}}  \\
    \midrule
     Gemini-3.1-Flash-Lite \cite{comanici2025gemini} & 69.6 & 81.5 / 74.5 & 81.2 & 93.8 / 68.6 & 40.0 & 62.5 & 27.5 & 46.6 & 70.1 & \textbf{65.08}\\ 
    +\methodname & 72.3 &82.9 / 75.1 & 83.3 & 94.6 / 70.8 & 51.5 & 69.0 & 36.0 & 57.8 & 82.7 & \textbf{70.55}\\
    \rowcolor{gray!10}
    \textit{Improvement} & \textcolor[RGB]{0, 150, 50}{+2.7} & 
    \textcolor[RGB]{0, 150, 50}{+1.4 / 0.6} &
    \textcolor[RGB]{0, 150, 50}{+2.1} &
    \textcolor[RGB]{0, 150, 50}{+0.8 / 2.2} &
    \textcolor[RGB]{0, 150, 50}{+11.5} &
    \textcolor[RGB]{0, 150, 50}{+6.5} &
    \textcolor[RGB]{0, 150, 50}{+8.5}&
    \textcolor[RGB]{0, 150, 50}{+11.2} &
    \textcolor[RGB]{0, 120, 50}{+12.6} & \textcolor[RGB]{0, 150, 50}{\textbf{+5.47}} \\
    \midrule
     Kimi K2.5 \cite{team2026kimi} & 85.9 & 85.6 / 80.6& 93.7 & 98.3 / 77.1 & 95.5 & 52.5 & 44.5 & 54.9 & 73.4 & \textbf{76.55}\\ 
    +\methodname & 87.4 & 86.8 / 81.1  & 94.1 & 98.4  / 77.9 & 98.0 & 57.0 & 70.5 & 57.7 & 83.9 & \textbf{81.16}\\
    \rowcolor{gray!10}
    \textit{Improvement} & \textcolor[RGB]{0, 150, 50}{+1.5} & 
    \textcolor[RGB]{0, 150, 50}{+1.2 / 0.5} &
    \textcolor[RGB]{0, 150, 50}{+0.4} &
    \textcolor[RGB]{0, 150, 50}{+0.1 / 0.8} &
    \textcolor[RGB]{0, 150, 50}{+2.5} &
    \textcolor[RGB]{0, 150, 50}{+4.5} &
    \textcolor[RGB]{0, 150, 50}{+26.0}&
    \textcolor[RGB]{0, 150, 50}{+2.8} &
    \textcolor[RGB]{0, 120, 50}{+10.5} & \textcolor[RGB]{0, 150, 50}{\textbf{+4.61}} \\
    \bottomrule
  \end{tabular}
  }
  \label{tab:main}
\end{table}

\begin{table}[!t]
  \begin{minipage}{0.48\textwidth}
  \centering
  \caption{\textbf{Comparison with a closed-source image editor.} Pass@1 of Qwen3-VL-8B and Gemini-3.1-Flash-Lite with Nano Banana 2 (NB2) vs.\ Ours on a 100-sample-per-benchmark subset.}
  \resizebox{\textwidth}{!}{
  \normalsize
  \begin{tabular}{l|c|c|c|c|c|c}
    \toprule
    Model & Per. & Cha. & Log. & Jig. &3D & \textbf{Avg.}\\
    \midrule
     Qwen3-VL \cite{bai2025qwen3} & 80.3 & 75.7 & 17.5 & 11.0 & 56.0 & 48.10 \\
     + NB2 \cite{google2026nano} & 82.0 & 77.0 & 20.5 & 13.0 & 55.5 & 49.60\\
     \textbf{+ Ours} & 82.0 & 77.3 & 24.0 &15.0 & 66.0 & \textbf{52.86}\\
     \midrule
     Gemini-3.1 \cite{comanici2025gemini} & 75.3 & 83.0 & 50.5 & 28.0 & 60.0 & 59.36\\
     + NB2 \cite{google2026nano} & 77.7  & 84.0 & 53.5 & 32.0 & 59.0 & 61.24\\
     \textbf{+ Ours} & 77.3 & 84.3 & 61.0 & 35.0 & 68.5 & \textbf{65.22}\\
    \bottomrule
  \end{tabular}
  }
  \label{tab:nano}
  \end{minipage}
  \hfil
  \begin{minipage}{0.48\textwidth}
    \caption{\textbf{Ablation of the reflection mechanism.} Pass@1 with and without the verification step in the Edit-Verify-Reason pipeline across Perception (Per.), Chart (Cha.), Logic (Log.), Jigsaw (Jig.), and 3D.}
  \centering
  \resizebox{\textwidth}{!}{
  \normalsize
  \begin{tabular}{l|c|c|c|c|c|c}
    \toprule
    Model & Per. & Cha. & Log. & Jig. & 3D & \textbf{Avg.}\\
    \midrule
     Qwen3-VL \cite{bai2025qwen3} & 79.1 & 73.2 & 18.5 & 9.5 & 56.0 & 47.26\\
     w.o. reflection & 79.3 & 73.2 & 25.5 & 13.0 & 68.2 & 51.84\\
     w. reflection & 80.9 & 75.0 & 26.0 & 13.0 & 67.9 & \textbf{52.56}\\
     \midrule
     Gemini-3.1 \cite{comanici2025gemini} & 75.2 & 81.2 & 51.3 & 27.5 & 58.4 &58.72\\
     w.o. reflection & 76.0 & 81.8 & 59.0 & 34.5 & 70.7 &64.40\\
     w. reflection & 76.8 & 82.9 & 60.3 & 36.0 & 70.3 & \textbf{65.26}\\
    \bottomrule
  \end{tabular}
  }
\label{tab:ablation_reflection}
\end{minipage}
\end{table}

\subsection{Main Results}
\textbf{Overall Performance.} Tab.~\ref{tab:main} compares \methodname against the two prevailing paradigms. Tool-based and programmatic models are competitive on chart understanding and within a few points on fine-grained perception, but their predefined action spaces do not cover tasks requiring non-local edits (Logic, Jigsaw, 3D), which we mark with $\dagger$ to indicate unsupported task scope rather than evaluated failure. Unified models avoid this coverage limit by natively interleaving generation, yet on average they trail specialized understanding backbones by more than ten points (e.g., ThinkMorph-7B \cite{gu2025thinkmorph} at $44.05$ and Bagel-Zebra-CoT \cite{li2025zebra} at $38.27$ vs.\ Qwen3-VL-8B \cite{bai2025qwen3} at $55.95$), reflecting the well-known tension between a jointly optimized generative head and pure understanding quality. \methodname sidesteps both pitfalls by treating a specialist editing model as a question-conditioned reasoning module, and improves every evaluated backbone (Qwen3-VL-8B \cite{bai2025qwen3}, Gemini-3.1-Flash-Lite \cite{comanici2025gemini}, Kimi K2.5 \cite{team2026kimi}) across the task families where comparisons are available.

\textbf{Comparison with a Closed-Source Editor.} Tab.~\ref{tab:nano} compares our editor against the closed-source Nano Banana 2~\cite{google2026nano} as the visual-thought provider for Qwen3-VL-8B \cite{bai2025qwen3} and Gemini-3.1-Flash-Lite \cite{comanici2025gemini}. To contain API cost, both editors are evaluated on the same 100-sample-per-benchmark subset, so absolute numbers should be read as trends rather than significance-tested estimates. Both editors improve over the no-edit baseline on most tasks, confirming that question-conditioned edits are useful even when produced by a generic frontier editor. Our editor is comparable to Nano Banana 2 on Perception and Chart and shows larger margins on Logic, Jigsaw, and 3D, where edits must encode task structure rather than re-render local regions. Results are consistent with our hypothesis that reasoning-aware training, not editor scale, drives guidance quality on structural tasks.

\subsection{Ablation Analysis}
To isolate each design choice in \methodname, we ablate (i) the two-stage training recipe (Tab.~\ref{tab:ablation}), (ii) the two RL reward signals, Editing Correctness and Editing Guidance (Tab.~\ref{tab:reward}), and (iii) the inference-time reflection mechanism (Tab.~\ref{tab:ablation_reflection}). All settings are evaluated with both understanding backbones, Qwen3-VL-8B \cite{bai2025qwen3} and Gemini-3.1-Flash-Lite \cite{comanici2025gemini}.

\textbf{Training Recipe:} Tab.~\ref{tab:ablation} compares the base FLUX.2-klein-base-9B \cite{blackforestlabs2025flux2} editor with its Stage~I SFT and Stage~II RL checkpoints. The base editor (where evaluable) tracks the no-edit backbone. Stage~I SFT improves all five task families on both backbones, suggesting the LLM encoder retains the reasoning capacity needed for question-conditioned editing and that supervised imitation suffices to reactivate it. Stage~II RL adds under one point on Perception and Chart and is essentially flat on Logic, Jigsaw, and 3D. We attribute this to the \emph{sampling granularity} of GRPO \cite{shao2024deepseekmath}: Perception and Chart involve localized edits whose group-sampled variants are spatially distinguishable by the reward, whereas Jigsaw and 3D require structural edits, which group sampling appears less able to produce, motivating future work on hierarchical or semantic-aware sampling.

\textbf{Reward Mechanism:} Tab.~\ref{tab:reward} isolates the two RL reward signals introduced in Sec.~\ref{sec:rl}: \emph{Editing Correctness} (isolated edit quality judged by a VLM-as-Judge \cite{zheng2023judging}) and \emph{Editing Guidance} (downstream answer accuracy). Each reward has complementary blind spots, and neither alone matches the full model. Guidance-only is on par with Correctness-only on Logic, where the downstream model benefits directly from edits that expose task-relevant structure, whereas Correctness-only is slightly stronger on Perception and Chart, where the judge can reliably assess localized visual edits such as bounding boxes and chart annotations. Combining the two matches or exceeds either reward alone across all task families: Correctness provides a fidelity floor by filtering implausible edits, while Guidance raises the task ceiling by ensuring each edit advances the reasoning chain.

\begin{table}[!t]
 \begin{minipage}{0.495\textwidth}
  \centering
  \caption{\textbf{Ablation of the two-stage training recipe.} Pass@1 of the base FLUX.2-klein-base-9B \cite{blackforestlabs2025flux2} editor, after Stage~I SFT (\textit{+sft}), and after Stage~II RL (\textit{+sft\&rl}), paired with Qwen3-VL-8B \cite{bai2025qwen3} and Gemini-3.1-Flash-Lite \cite{comanici2025gemini}.}
  \resizebox{\textwidth}{!}{
  \normalsize
  \begin{tabular}{l|c|c|c|c|c|c}
    \toprule
    Model & Per. & Cha. & Log. & Jig. &3D & \textbf{Avg.} \\
    \midrule
     Qwen3-VL \cite{bai2025qwen3} & 79.1 & 73.2 & 18.5 & 9.5 & 56.0 & 47.26 \\
     \midrule
     Flux \cite{blackforestlabs2025flux2} & 74.3 & 68.9 & 18.5 & 9.5 & 55.1 & 45.26\\
     + sft & 79.7 & 74.1 & 25.8& 13.0 & 67.3 & 51.98\\
     + sft\&rl & 80.9 & 75.0 & 26.0 & 13.0 & 67.9 & \textbf{52.56}\\
     \midrule
     Gemini-3.1 \cite{comanici2025gemini} & 75.2 & 81.2 & 51.3 & 27.5 & 58.4 & 58.72\\
     \midrule
     Flux \cite{blackforestlabs2025flux2} & 70.3 & 74.5 & 51.3 & 27.0 & 59.5 & 56.52\\
     + sft & 76.0 & 82.4 & 59.5 & 36.0 & 70.0 & 64.78\\
     + sft\&rl & 76.8 & 82.9 & 60.3 & 36.0 & 70.3 & \textbf{65.26}\\
    \bottomrule
  \end{tabular}
  }
  \label{tab:ablation}
  \end{minipage} 
\hfil
\begin{minipage}{0.47\textwidth}
  \caption{\textbf{Ablation of the two RL reward signals in Stage~II RL.} Pass@1 with Editing Correctness only (Cor.), Editing Guidance only (Gui.), or both, , paired with Qwen3-VL-8B.}
  \centering
  \resizebox{\textwidth}{!}{
  \normalsize
  \begin{tabular}{c|c|c|c|c|c|c|c}
    \toprule
    Cor. & Gui. & Per. & Cha. & Log. & Jig. & 3D &\textbf{Avg.}\\
    \midrule
    \rowcolor{blue!8}
    \multicolumn{8}{c}{\textit{Qwen3-VL-8B-Instruct} \cite{bai2025qwen3}} \\
    \cmidrule(){1-8}
    \Checkmark &\XSolidBrush & 80.7 & 74.7 & 25.8 & 13.0 & 66.8 & 52.20\\
    \XSolidBrush &\Checkmark & 80.2 & 74.2 & 26.0 & 13.0 & 67.6 &52.20\\
    \Checkmark &\Checkmark & 80.9 & 75.0 & 26.0 & 13.0 & 67.9 &\textbf{52.56}\\
    \midrule
    \rowcolor{blue!8}
    \multicolumn{8}{c}{\textit{Gemini-3.1-flash-lite-preview} \cite{comanici2025gemini}} \\
    \cmidrule(){1-8}
    \Checkmark &\XSolidBrush  & 76.6 & 82.5 & 60.0 & 35.5 & 69.7 & 64.86\\
    \XSolidBrush &\Checkmark &  76.1 & 82.5 & 60.0 & 36.0 & 69.9 &64.90 \\
    \Checkmark &\Checkmark & 76.8 & 82.9 & 60.3 & 36.0 & 70.3 & \textbf{65.26}\\
    \bottomrule
  \end{tabular}
  }
  \label{tab:reward}
\end{minipage}
\end{table}

\textbf{Reflection Mechanism.} Tab.~\ref{tab:ablation_reflection} ablates the Edit-Verify-Reason pipeline by removing the verification step, so every edited image is passed to the understanding model. Reflection yields consistent gains on Perception and Chart, where the judge can reliably flag mis-localized edits and where backbone accuracy is already high (around 80\%); in this regime, suppressing harmful edits matters more than supplying additional hints. Gains are smaller on Logic and absent on Jigsaw for Qwen3-VL-8B \cite{bai2025qwen3}, with a small regression on 3D. Baseline accuracy is low on these tasks. The model rarely solves these tasks from the original image alone, and we hypothesize that even imperfect edits are more useful as hints than discarded. This suggests reflection should be applied selectively, gated by backbone confidence or task family rather than uniformly.

\begin{figure}[t]
  \centering
    \includegraphics[width=\linewidth]{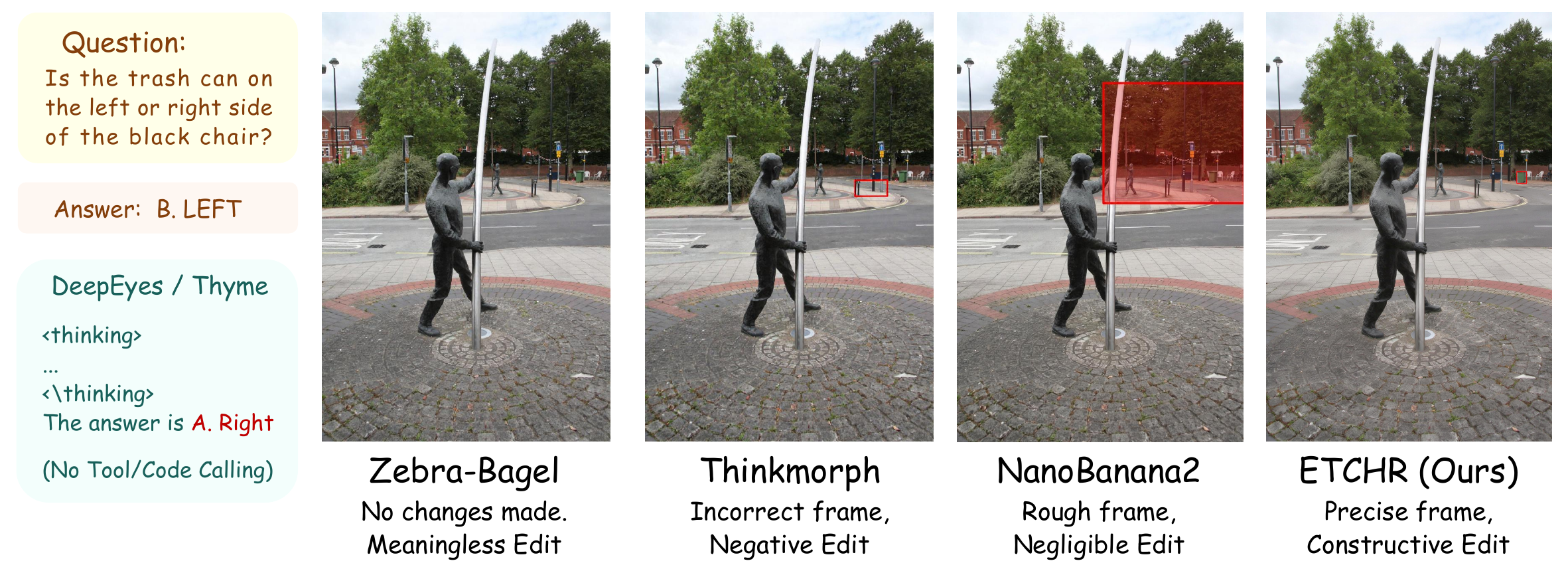} 
    \caption{\textbf{Qualitative comparison} of \methodname with existing methods. \methodname provides the most precise guidance for this problem, assisting the reasoning model in generating the correct answer.}
    \label{fig:case}
\end{figure}

\noindent \textbf{Case Analysis.}
Fig.~\ref{fig:case} demonstrates how various 'think with images' strategies assist comprehension through image editing. DeepEyesV2 and Thyme failed to invoke the editing tools, leading to incorrect final answers. Bagel and ThinkMorph performed incorrect edits, also resulting in wrong answers. While Nano Banana 2 executed the edit correctly, the highlighted regions lacked precision, providing minimal guidance for the reasoning model. In contrast, our editing model provides extremely precise edits, enabling the reasoning model to answer the question correctly.

\section{Related Work}

\noindent \textbf{Thinking with Images: Tool-based Pipelines.} Tool-based ``think with images'' methods couple the understanding MLLM with a deterministic external renderer: the MLLM emits a structured action and the renderer executes it on the input image, producing the intermediate visual aid that feeds the next reasoning step. Visual Sketchpad~\cite{hu2024visual} is an early instance, prompting the MLLM to draw spatial annotations and auxiliary lines for geometric and mathematical reasoning. V*~\cite{wu2024v} and DeepEyes~\cite{zheng2025deepeyes} specialize this idea to fine-grained visual search by emitting bounding-box and crop-zoom commands that progressively localize the queried region, while Thyme~\cite{zhang2025thyme} and DeepEyes-V2~\cite{hong2025deepeyesv2} extend the action space to executable code snippets for chart parsing and multi-step chart understanding. Although the resulting actions are interpretable and reproducible, two limitations persist. First, the action space is fixed at design time, so global, free-form, or task-novel transformations (e.g., redrawing a maze path, reordering jigsaw tiles, re-rendering a 3D viewpoint) lie outside what the renderer can produce. Second, each new family of actions typically requires task-specific fine-tuning of the understanding model, which can erode its general competence~\cite{kirkpatrick2017overcoming} and forces a separate adaptation per deployment. \methodname avoids both costs by replacing the deterministic renderer with a question-conditioned image editor whose transformation space is continuous, and by keeping the understanding model frozen so a single trained editor plugs into different open- and closed-source MLLMs without further fine-tuning.

\noindent \textbf{Thinking with Images: Unified Multi-modal Models.} A second family pursues ``think with images'' with a single backbone that natively interleaves text and image tokens, removing the external renderer altogether. Chameleon~\cite{team2024chameleon} and Show-o~\cite{xie2024show} demonstrate joint autoregressive and diffusion modeling over a shared token space, while Janus~\cite{wu2025janus} decouples the visual encoder for understanding from the one used for generation to relieve the optimization tension between the two objectives. More recently, ThinkMorph~\cite{gu2025thinkmorph} and Zebra-CoT~\cite{li2025zebra} fine-tune unified models on interleaved reasoning traces and report emergent adaptive modality switching. Despite this flexibility, the jointly optimized generative head trails specialist image editors in fidelity, so its intermediate images frequently inject noise rather than evidence, as quantified by UniG2U-Bench~\cite{wen2026unig2u}; the same backbone also lags specialist understanding models on pure perception tasks~\cite{wu2025janus,xie2024show}. \methodname sidesteps this failure by keeping understanding and generation in separate specialist models: the editor is trained on reasoning-aware objectives without touching the understanding MLLM, and edit fidelity is preserved because the editor is not asked to also serve as the answer producer.

\noindent \textbf{Image Editing Models.} Recent image-to-image editing methods such as FLUX-class models~\cite{blackforestlabs2025flux2} and Qwen-Image-Edit~\cite{wu2025qwen} replace shallow CLIP-style text encoders with MLLM-style encoders, expanding the editor's capacity to parse complex instructions. Closest in spirit are instruction-tuned diffusion editors: InstructPix2Pix~\cite{brooks2023instructpix2pix} fine-tunes a latent diffusion model on synthetic before/after pairs, and MagicBrush~\cite{zhang2023magicbrush} improves edit faithfulness with human-annotated real-image edits. Across all of these, however, the training objective remains anchored in aesthetic fidelity and instruction-following accuracy: the user is assumed to supply an explicit edit prompt, so the editor is optimized as a faithful instruction renderer and is never asked to infer what edit would help answer an abstract question. \methodname targets this gap by repurposing the MLLM-encoder editor as a question-conditioned, reasoning-aware module rather than a passive instruction-following renderer.
\section{Conclusion}
We present \methodname, a question-conditioned image editor for think with images reasoning. Our analysis identifies two gaps in previous works, a \emph{language-side} gap in mapping abstract questions to edits and a \emph{generation-side} gap under increasing reasoning depth, and motivates a two-stage recipe (Reasoning Imitation SFT and Reasoning Enhancement RL) paired with an Edit-Verify-Reason inference procedure. Decoupling the editor from the understanding model lets \methodname plug into open- and closed-source MLLMs in a training-free manner.

\textbf{Limitations:} \emph{GRPO sampling granularity:} GRPO’s  sampling exhibits limited semantic diversity for structural editing (e.g., jigsaw restoration), which constrains policy exploration and prevents uniform RL gains across all reasoning families. \emph{Downstream capability bound:} \methodname end-to-end performance remains capped by the understanding model; even optimally edited intermediates cannot overcome the reasoning ceiling of the downstream MLLM. \emph{Additional time cost:} Image editing incurs a higher temporal cost compared to text-based reasoning; consequently, \methodname increases the time overhead for samples requiring only a short CoT.

\bibliographystyle{plain}
\bibliography{refs}

\clearpage
\appendix
\section{Prompts}
\textbf{Task-level Prompt:}

\emph{Fine-grained Perception:} Draw a red box to mark the important regions for this question in the figure.

\emph{Chart Understanding:}
Draw a red box to mark the important regions for this question in the figure.

\emph{Logic:} Draw the shortest path of the maze in blue.

\emph{Jigsaw:} Draw the original image after restoring this jigsaw puzzle task.

\emph{3D Understanding: } Imagine a new perspective of the original image that helps answer the question.

\noindent \textbf{Verify Prompt:}

\emph{Fine-grained Perception:} Please judge whether the information relevant to solving this problem is marked with a red box in the picture. If the red box contains valid information, reply with 1; if there is no red box in the picture, or the area enclosed by the red box does not contain valid information, reply with 0.

\emph{Chart Understanding:}
Please judge whether the information relevant to solving this problem is marked with a red box in the picture. If the red box contains valid information, reply with 1; if there is no red box in the picture, or the area enclosed by the red box does not contain valid information, reply with 0.

\emph{Maze:} Please determine whether the following image contains a valid maze path: the middle path (blue) connects the starting point (green) to the destination (red), is continuous in four directions (up, down, left, right) with no diagonal movement allowed, and no black obstacles or white path tiles have been altered compared to Figure 1 (the original maze) in Figure 2 (the path diagram). If the maze if valid, reply with 1. Otherwise, return with 0.

\emph{Frozen Lake:} Please determine whether the path drawn with red lines in Figure 2  connects the treasure to the character without passing through any holes. If the frozen lake if valid, reply with 1. Otherwise, reply with 0.

\emph{Jigsaw:} Please determine whether Figure 2 is a correct restoration of the jigsaw puzzle task in Figure 1. If Fig.2 is correct, reply with 1. Otherwise, reply with 0.

\emph{3D Understanding: } judge: Please judge whether you can see the objects mentioned in the question from the perspective of the second image. If the objects are presented in the image, reply with 1. Otherwise, reply with 0.

\noindent \textbf{Reason Prompt:} 

\emph{Fine-grained Perception:} Answer the question based on the two images above. The first image is the original one corresponding to the question, while the second one uses a red box to mark the area containing the key information for the question in the image. Please first focus on the boxed part in the second image, and answer the question based on the corresponding information in the two images.

\emph{Chart Understanding:}
Answer the question based on the two images above. The first image is the original one corresponding to the question, while the second one uses a red box to mark the area containing the key information for the question in the image. Please first focus on the boxed part in the second image, and answer the question based on the corresponding information in the two images.

\emph{Maze:} Solve the maze problem in the first image. The blue path in the second image is very likely the correct solution to the maze.

\emph{Frozen Lake:} Solve the frozen lake problem and output the path from the character to the treasure. The blue path in the second image is very likely the correct solution to the frozen lake problem.

\emph{Jigsaw:} Restore the jigsaw image based on the two images. The second image is likely the correct restoration of the jigsaw puzzle task from the first image.

\emph{3D Understanding: } Answer the question based on the two images above. The first image is the original one corresponding to the question, while the second one provides a novel perspective to help you solve the problem. Please focus on the second image to answer the question.

\section{More Cases}
\begin{figure}[t]
  \centering
    \includegraphics[width=\linewidth]{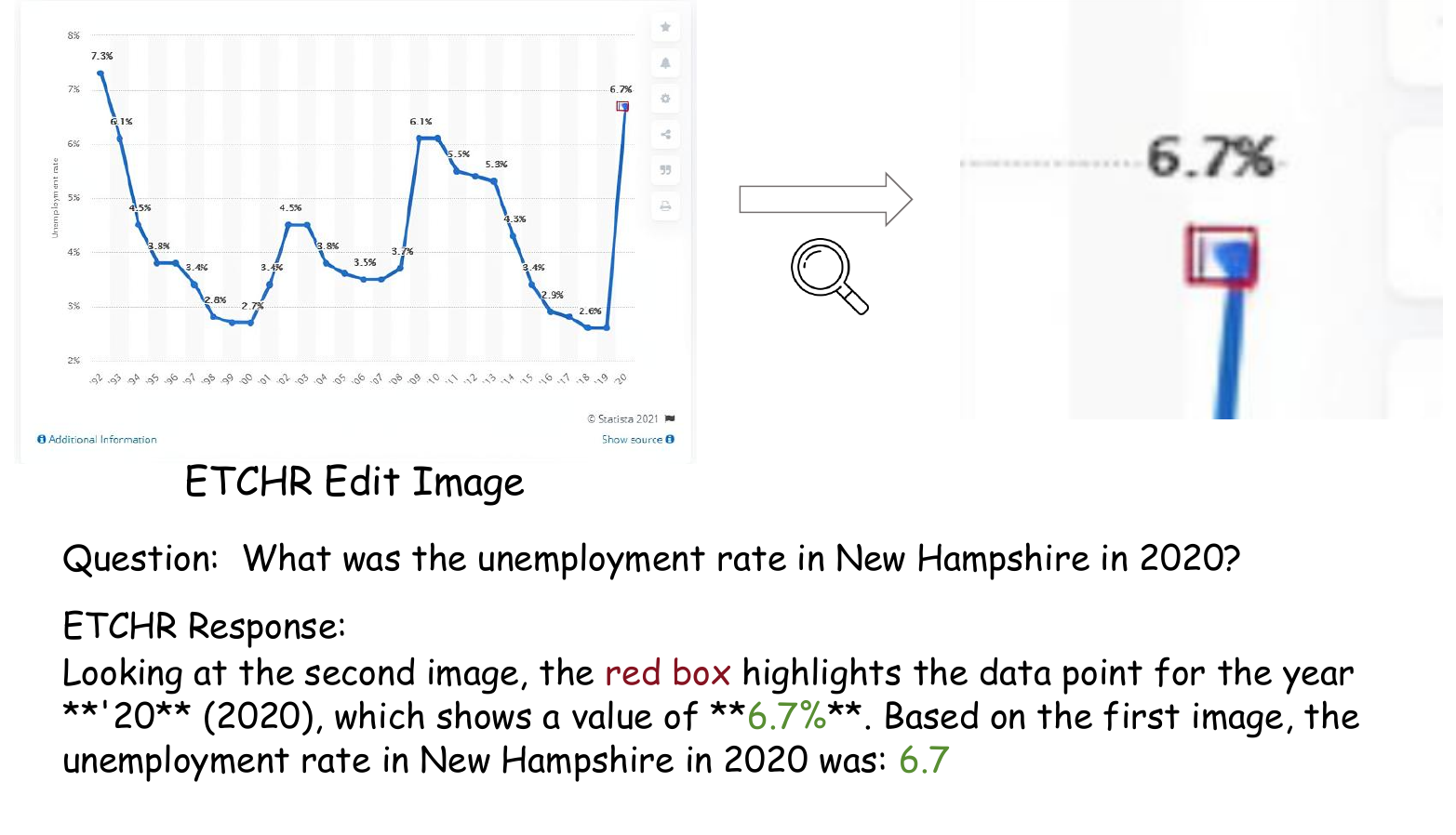}
    \caption{\methodname enables precise localization of granular information in charts, facilitating a better comprehension of the visual data for understanding models.}
    \label{fig:case-chart}
\end{figure}

\begin{figure}[t]
  \centering
    \includegraphics[width=\linewidth]{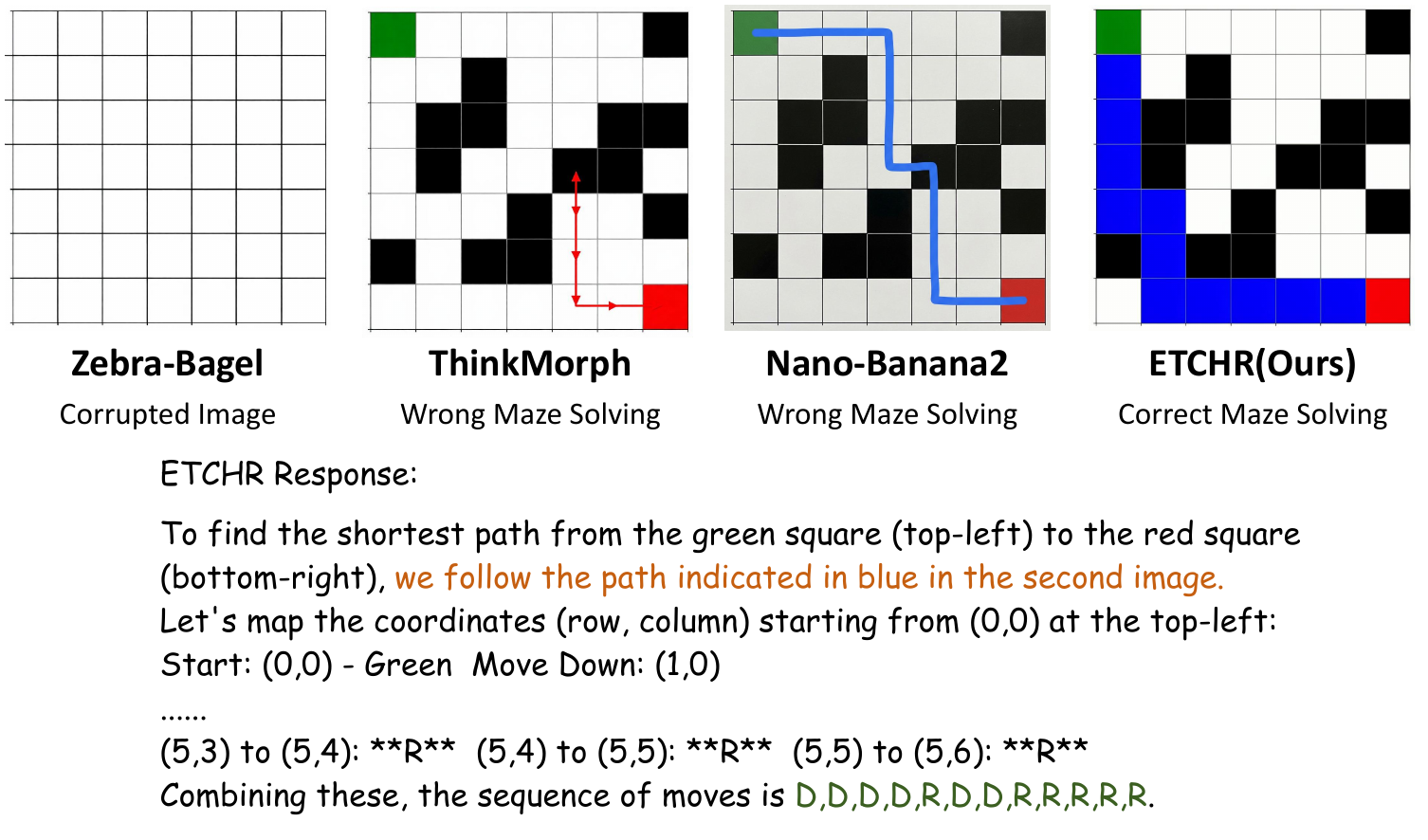}
    \caption{\methodname can draw the correct path on the maze map, helping understanding model accurately identify the route to exit the maze.}
    \label{fig:case-maze}
\end{figure}

We further provide examples of \methodname assisting models in performing reasoning across tasks such as chart understanding (Fig.~\ref{fig:case-chart}), logical reasoning (Fig.~\ref{fig:case-maze}), jigsaw puzzles, and 3D perception (Fig.~\ref{fig:case-3D}).

\begin{figure}[t]
  \centering
    \includegraphics[width=\linewidth]{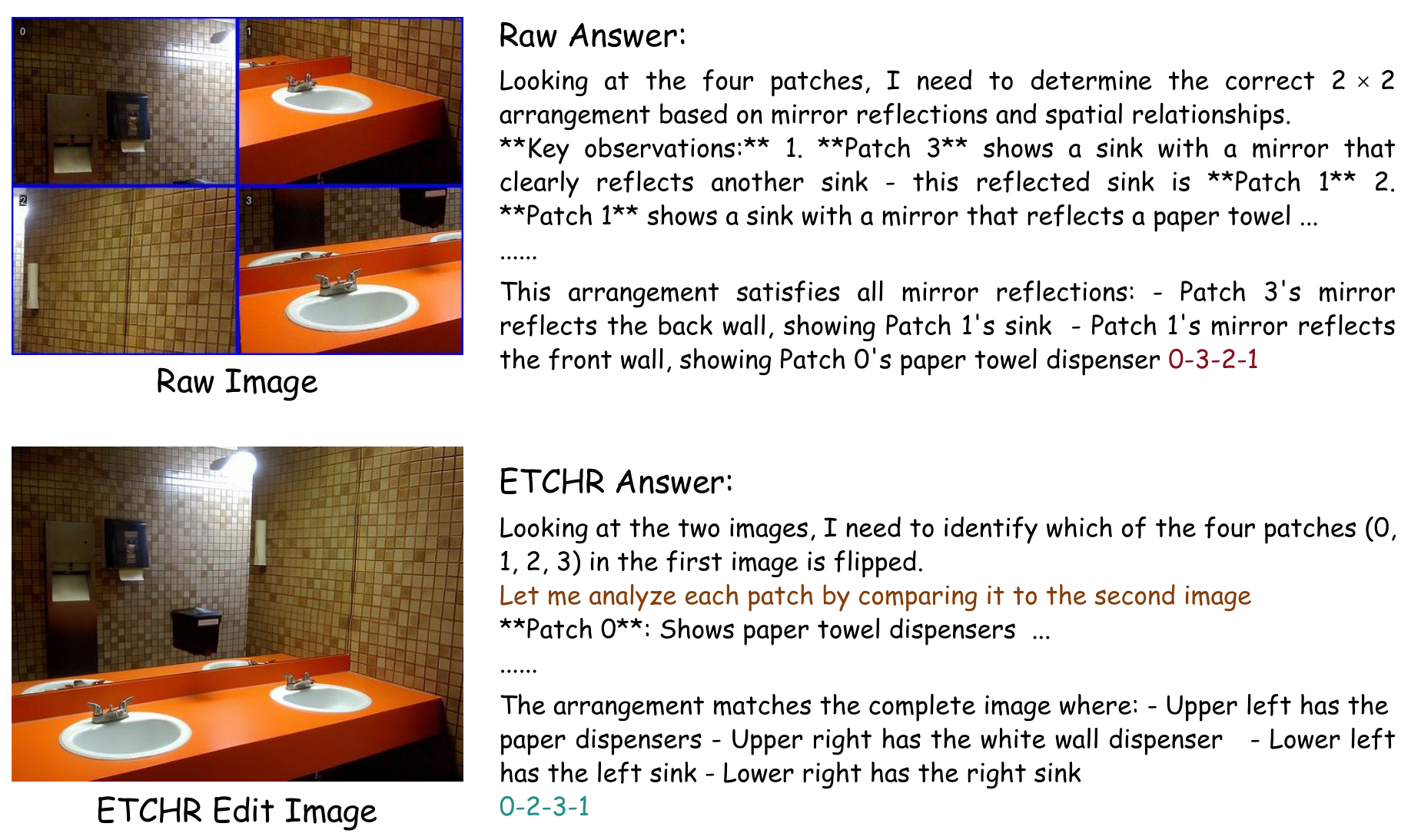}
    \caption{\methodname is capable of restoring scrambled jigsaw images, thereby guiding the understanding model to correctly solve jigsaw tasks.}
    \label{fig:case-jigsaw}
\end{figure}

\begin{figure}[t]
  \centering
    \includegraphics[width=\linewidth]{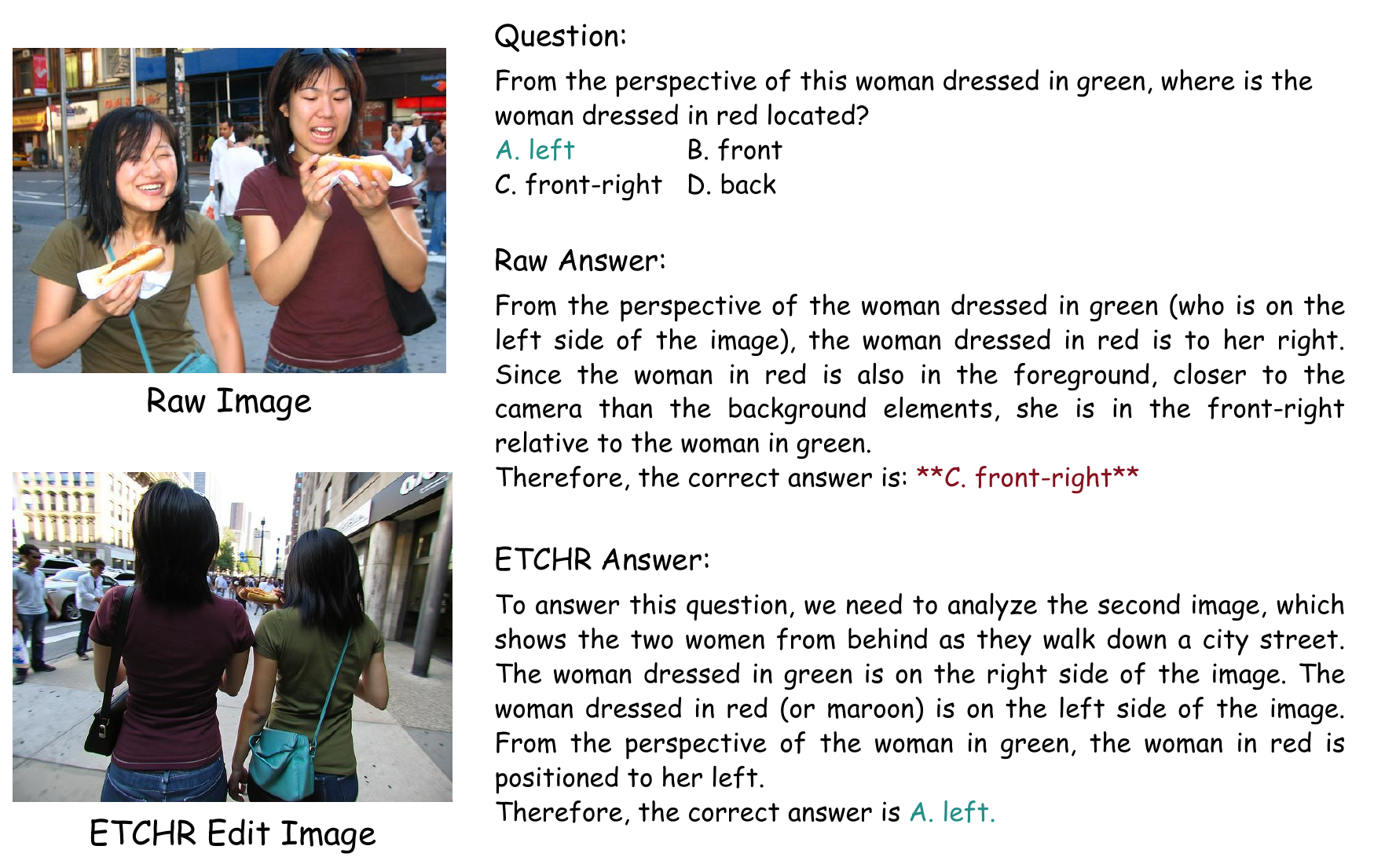}
    \caption{\methodname enables perspective shifting based on the specific question, helping to ensure the model correctly understands spatial positioning.}
    \label{fig:case-3D}
\end{figure}

\end{document}